\documentclass[runningheads]{llncs}
\usepackage{graphicx}
\usepackage{xcolor}
\usepackage{amsmath, amsfonts, amssymb}
\usepackage{nicefrac}
\usepackage{multirow}
\DeclareMathOperator{\SE}{SE}

\begin{document}
\title{Generalizing Spatial Transformers to Projective Geometry with Applications to 2D/3D Registration\thanks{Supported by organization x.}}
\titlerunning{Generalizing Spatial Transformers to Projective Geometry}
%



\author{Cong Gao \and
Xingtong Liu \and
Wenhao Gu \and 
Benjamin Killeen \and 
Mehran Armand \and
Russell Taylor \and
Mathias Unberath}
\authorrunning{C. Gao et al.}
%
\institute{Johns Hopkins University, Baltimore MD 21218, USA\\
\email{cgao11@jhu.edu}}
\maketitle              
\begin{abstract}
Differentiable rendering is a technique to connect 3D scenes with corresponding 2D images. Since it is differentiable, processes during image formation can be learned. Previous approaches to differentiable rendering focus on mesh-based representations of 3D scenes, which is inappropriate for medical applications where volumetric, voxelized models are used to represent anatomy. 
We propose a novel Projective Spatial Transformer module that generalizes spatial transformers to projective geometry, thus enabling differentiable volume rendering. We demonstrate the usefulness of this architecture on the example of 2D/3D registration between radiographs and CT scans. Specifically, we show that our transformer enables end-to-end learning of an image processing and projection model that approximates an image similarity function that is convex with respect to the pose parameters, and can thus be optimized effectively using conventional gradient descent. To the best of our knowledge, this is the first time that spatial transformers have been described for projective geometry. The source code will be made public upon publication of this manuscript and we hope that our developments will benefit related 3D research applications. 

\end{abstract}
%
%
\section{Introduction}
Differentiable renderers that connect 3D scenes with 2D images thereof have recently received considerable attention~\cite{loper2014opendr,henderson2019learning,liu2019soft} as they allow for simulating, and more importantly \emph{inverting}, the physical process of image formation. Such approaches are designed for integration with gradient-based machine learning techniques including deep learning to, e.g., enable single-view 3D scene reconstruction. Previous approaches to differentiable rendering have largely focused on mesh-based representation of 3D scenes. This is because compared to say, volumetric representations, mesh parameterizations provide a good compromise between spatial resolution and data volume. 
Unfortunately, for most medical applications the 3D scene of interest, namely the anatomy, is acquired in volumetric representation where every voxel represents some specific physical property. Deriving mesh-based representations of anatomy from volumetric data is possible in some cases~\cite{gibson2018automatic}, but is not yet feasible nor desirable in general, since surface representations cannot account for tissue variations within one closed surface. However, solutions to the differentiable rendering problem are particularly desirable for X-ray-based imaging modalities, where 3D content is reconstructed from -- or aligned to multiple 2D transmission images. This latter process is commonly referred to as 2D/3D registration and we will use it as a test-bed within this manuscript to demonstrate the value of our method.

Mathematically, the mapping from volumetric 3D scene $V$ to projective transmission image $I_m$ can be modeled as $I_m=A(\theta)V$, where $A(\theta)$ is the system matrix that depends on pose parameter $\theta \in \SE(3)$. 
In intensity-based 2D/3D registration, we seek to retrieve the pose parameter $\theta$ such that the image $I_m$ simulated from the 3D CT scan $V$ is as similar as possible to the acquired image $I_f$:
\begin{equation}
    \min_{\theta}L(I_f, I_m)\\
    = \min_{\theta}L(I_f, A(\theta)V),
\end{equation}
where $L$ is the similarity function. Gradient decent-based optimization methods require the gradient $\frac{\partial \mathcal{L}}{\partial \mathbf{\theta}} = \frac{\partial \mathcal{L}}{\partial A(\mathbf{\theta})}\cdot \frac{\partial A(\mathbf{\theta})}{\partial \mathbf{\theta}}$
at every iteration. Although the mapping was constructed to be differentiable, analytic gradient computation is still impossible due to excessively large memory footprint of $A$ for all practical problem sizes\footnote{It is worth mentioning that this problem can be circumvented via ray casting-based implementations if one is interested in $\nicefrac{\partial L}{\partial V}$ but not in $\nicefrac{\partial L}{\partial \theta}$~\cite{wurfl2018deep}.}. This prevents the use of volumetric rendering in gradient-based machine learning techniques.
In this work, we propose an analytically differentiable volume renderer that follows the terminology of spatial transformer networks~\cite{jaderberg2015spatial} and extends their capabilities to spatial transformations in projective geometry. Our specific contributions are:
\begin{itemize}
\item We introduce a \textbf{Pro}jective \textbf{S}patial \textbf{T}ransformer~(ProST) module that generalizes spatial transformers~\cite{jaderberg2015spatial} to projective geometry. This enables volumetric rendering of transmission images that is differentiable both with respect to the input volume $\mathbf{x}$ as well as the pose parameters $\mathbf{\theta}$.
\item We demonstrate how ProST can be used to solve the non-convexity problem of conventional intensity-based 2D/3D registration. Specifically, we train an end-to-end deep learning model to approximate a convex loss function derived from geodesic distances between poses $\mathbf{\theta}$ and enforce desirable pose updates $\frac{\partial L}{\partial \theta}$ via double backward functions on the computational graph.
\end{itemize}

\section{Methodology}
\subsection{Projective Spatial Transformer~(ProST)}
\begin{figure}[h!]
\includegraphics[width=\textwidth]{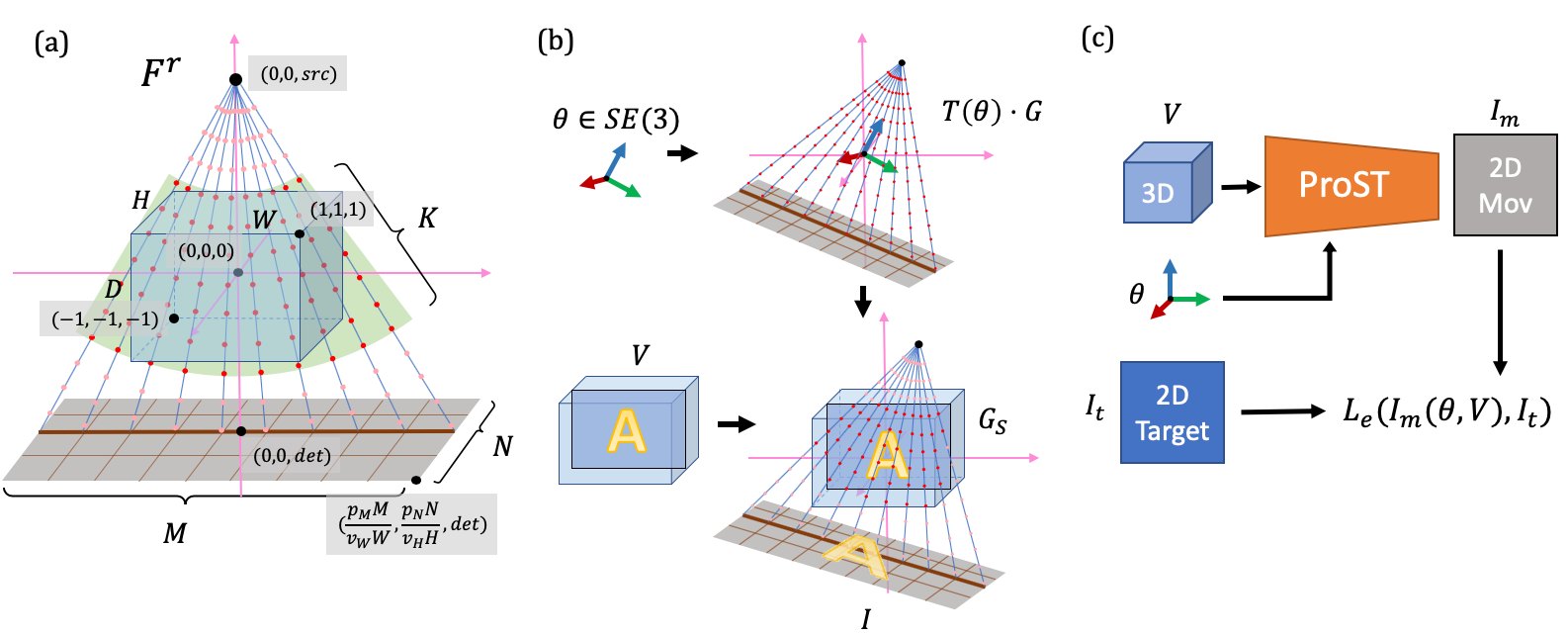}
\caption{(a) Canonical projection geometry and a slice of cone-beam grid points are presented with key annotations. The green fan covers the control points which are used for further reshape. (b) Illustration of grid sampling transformer and projection. (c) Scheme of applying ProST to 2D/3D registration.} \label{fig1:geometry}
\end{figure}

\subsubsection{Canonical projection geometry}
Given a volume $V \in \mathbb{R}^{D \times W \times H}$ with voxel size $v_D \times v_W \times v_H$, we define a reference frame $F^r$ with the origin at the center of $V$.
We use normalized coordinates for depth~($Dv_D$), width~($Wv_W$) and height~($Hv_H$), so that the points of $V$ are contained within the unit cube $(d, w, h) \in [-1, 1]^3$.
Given a camera intrinsic matrix $\mathcal{K} \in \mathbb{R}^{3 \times 3}$, we denote the associated source point as $(0, 0, src)$ in $F^r$.
The spatial grid $G$ of control points, shown in Fig.~\ref{fig1:geometry}-(a), lies on $M\times N$ rays originating from this source.
Because the control points in regions where no CT voxels exist will not contribute to the line integral, we cut the grid point cloud to a cone-shape structure that covers the exact volume space.
Thus, each ray has $K$ control points uniformly spaced within the volume $V$, so that the matrix $G \in \mathbb{R}^{4 \times (M\cdot N\cdot K)}$ of control points is well-defined, where each column is a control point in homogeneous coordinates.
These rays describe a cone-beam geometry which intersects with the detection plane, centered on $(0, 0, det)$ and perpendicular to the $z$ axis, as determined by $\mathcal{K}$.
The upper-right corner of the detection plane is at $(\frac{p_MM}{v_WW},\frac{p_NN}{v_HH},det)$.




\subsubsection{Grid sampling transformer} Our Projective Spatial Transformer (ProST) extends the canonical projection geometry by learning a transformation of the control points $G$.
Given $\theta \in \SE(3)$, we obtain a transformed set of control points via the affine transformation matrix $T(\theta)$:
\begin{equation}
\label{eqn1}
    G_T = T(\theta)\cdot G,
\end{equation}
as well as source point $T(\theta) \cdot (0, 0, src, 1)$ and center of detection plane $T(\theta) \cdot (0, 0, det, 1)$.
Since these control points lie within the volume $V$ but in between voxels, we interpolate the values $G_S$ of $V$ at the control points.
\begin{equation}
\label{eqn2}
    G_S = \texttt{interp}(V, G_T),
\end{equation}
where $G_S \in \mathbb{R}^{M \times N \times K}$.
Finally, we obtain a 2D image $I\in \mathbb{R}^{M \times N}$ by integrating along each ray. This is accomplished by ``collapsing'' the $k$ dimension of $G_S$:
\begin{equation}
\label{eqn3}
    I^{(m,n)} = \sum_{k=1}^{K} G_S^{(m,n,k)}
\end{equation}


The process above takes advantage of the spatial transformer grid, which reduces the projection operation to a series of linear transformations. 
The intermediate variables are reasonably sized for modern computational graphics cards, and thus can be loaded as a tensor variable. We implement the grid generation function using the C++ and CUDA extension of the PyTorch framework and embed the projection operation as a PyTorch layer with tensor variables. With the help of PyTorch autograd function, this projection layer enables analytical gradient flow from the projection domain back to the spatial domain. Fig.~\ref{fig1:geometry}~(c) shows how this scheme is applyied to 2D/3D registration. With out any learning parameters, we can perform registration with PyTorch's powerful built-in optimizers on large-scale volume representations. 
Furthermore, integrating deep convolutional layers, we show that ProST makes end-to-end 2D/3D registration feasible.

\subsection{Approximating Convex Image Similarity Metrics}
Following \cite{grupp2019automatic}, we formulate an intensity-based 2D/3D registration problem with a pre-operative CT volume~$V$, Digitally Reconstructed Radiograph~(DRR) projection operator~$P$, pose parameter~$\theta$, a fixed target image~$I_f$, and a similarity metric loss~$L_S$:
\begin{equation}
    \min_{\theta \in SE(3)}L_S\Big(I_f, P(V; \theta)\Big).
\end{equation}
Using our projection layer $P$, we propose an end-to-end deep neural network architecture which will learn a convex similarity metric, aiming to extend the capture range of the initialization for 2D/3D registration. Geodesic loss, $L_G$, which is the square of geodesic distance in $SE(3)$, has been studied for registration problems due to its convexity~\cite{salehi2018real}~\cite{mahendran20173d}. 
We take the implementation of~\cite{miolane2018geomstats} to calculate the geodesic gradient $\frac{\partial L_G(\theta, \theta_t)}{\theta}$, given a sampling pose $\theta$ and a target pose $\theta_t$. We then use this geodesic gradient to train our network, making our training objective exactly the same as our target task -- learning a convex shape similarity metric.\\ 

\begin{figure}[h!]
\includegraphics[width=0.9\textwidth]{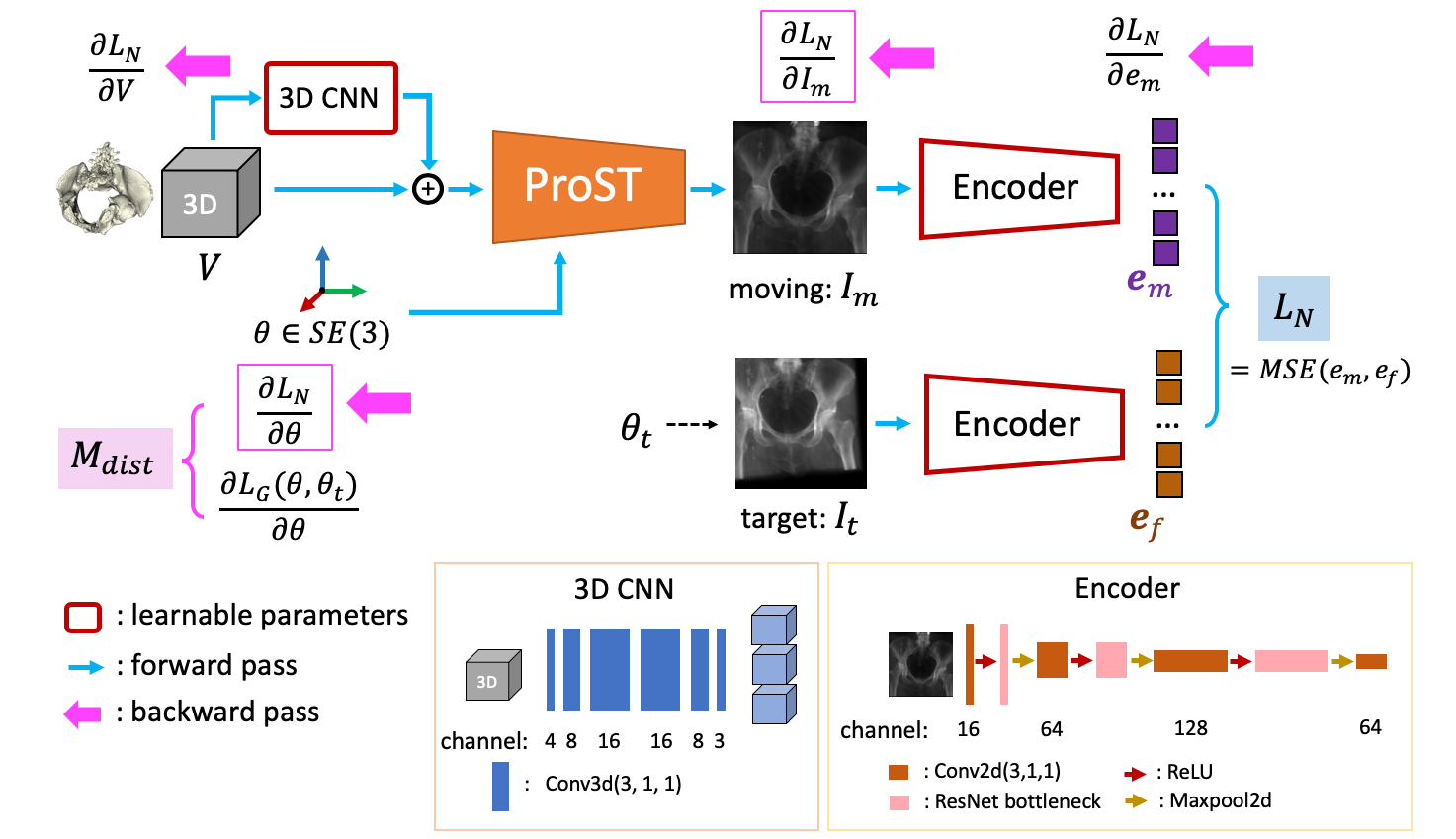}
\caption{DeepNet Architecture. Forward pass follows the blue arrows. Backward pass follows pink arrows, where gradient input and output of ProST in Eq.~\ref{eqn_grad} are highlighted with pink border.} \label{fig2:deepnet}
\end{figure}

Fig.~\ref{fig2:deepnet} shows our architecture. The input includes a 3D volume:~$V$, a pose parameter:~$\theta \in SE(3)$ and a target image:~$I_t$. All blocks which contain learnable parameters are highlighted with a red outline. The 3D CNN is a skip connection from the input volume to multi-channel expansion just to learn the residual. Projections are performed by projection layer with respect to $\theta$, which does not have learnable parameters. The projected moving image $I_m$ and the input target image $I_t$ go through two encoders, which are the same in structure but the weights are not shared, and output embedded features $e_m$ and $e_f$. Our network similarity metric $L_N$ is the mean squared error of $e_m$ and $e_f$. We will then explain the design from training phase and application phase separately.

\subsubsection{Training phase} The goal of training is to make the gradient of our network error function w.r.t. pose parameter, $\frac{\partial L_N}{\partial \theta}$,
close to the geodesic gradient $\frac{\partial L_G}{\partial \theta}$. The blue arrows in Fig.~\ref{fig2:deepnet} show the forward pass in a single iteration. The output can be written as $L_N(\phi;\theta, V, I_t)$, where $\phi$ are the network parameters. We then apply back-propagation, illustrated with pink arrows in Fig.~\ref{fig2:deepnet}. This yields $\frac{\partial L_N}{\partial \theta}$ and $\frac{\partial L_N}{\partial \phi}$. 
Assuming $L_N$ is the training loss, $\phi$ would normally be updated according to $lr\cdot \frac{\partial L_N}{\partial \phi}$.
However, we \textit{do not} update the network parameters during the backward pass. Instead we obtain the gradient and calculate a distance measure of these two gradient vectors, $M_{dist}(\frac{\partial L_N}{\partial \mathbf{\theta}}, \frac{\partial L_G}{\partial \mathbf{\theta}})$, which is our true network loss function during training. We perform a second forward pass, or ``double backward'' pass, to get $\frac{\partial M_{dist}}{\partial \phi}$ for updating network parameters $\phi$. To this end, we formulate the network training as the following optimization problem
\begin{equation}
    \min_{\phi}M_{dist}\Big(\frac{\partial L_N(\phi;V,\mathbf{\theta},I_f)}{\partial \mathbf{\theta}}, \frac{\partial L_G(\theta, \theta_t)}{\partial \mathbf{\theta}}\Big).
\end{equation}
Since the gradient direction is the most important during iteration in application phase, we design $M_{dist}$ by punishing the directional difference of these two gradient vectors. Translation and rotation are formulated using Eq.~\ref{eqn9}-\ref{eqn11}
\begin{equation}
    \label{eqn9}
    v^t_1, v^r_1=\Big(\frac{\partial L_N(\phi;V,\theta,I_f)}{\partial \mathbf{\theta}}\Big)_{trans, rot}; v^t_2, v^r_2=\Big(\frac{\partial L_G(\theta, \theta_t)}{\partial \mathbf{\theta}}\Big)_{trans, rot}
\end{equation}
\begin{equation}
    \label{eqn10}
    M_{dist}^{trans} = ||\frac{v_1^t}{||v_1^t||} - \frac{v_2^t}{||v_2^t||}||^2, M_{dist}^{rot} = ||\frac{v_1^r}{||v_1^r||} - \frac{v_2^r}{||v_2^r||}||^2
\end{equation}
\begin{equation}
    \label{eqn11}
    M_{dist} = M_{trans}+M_{rot},
\end{equation}
where the rotation vector is transformed into Rodrigues angle axis.

\subsubsection{Application phase} During registration, we fix the network parameters $\phi$ and start with an initial pose $\theta$. We can perform gradient-based optimization over $\theta$ based on the following back-propagation gradient flow
\begin{equation}
    \label{eqn_grad}
    \frac{\partial L_N}{\partial \theta}
    =\frac{\partial L_N}{\partial I_m}
    \cdot\frac{\partial I_m}{\partial G_S}
    \cdot\frac{\partial G_S}{\partial G_T}
    \cdot\frac{\partial G_T}{\partial T(\theta)}
    \cdot\frac{\partial T(\theta)}{\partial \theta}.
\end{equation}
The network similarity is more effective when the initial pose is far away from the groundtruth, while less senstive to local textures compared to traditional image-based methods, such as Gradient-based Normalized Corss Correlation~(Grad-NCC)~\cite{penney1998comparison}. We implement Grad-NCC as a pytorch loss function~$L_{GNCC}$, and combine these two methods to build an end-to-end pipeline for 2D/3D registration. We first detect the convergence of the network-based optimization process by monitoring the standard deviation~(STD) of $L_N$. After it converges, we then switch to optimize over $L_{GNCC}$ until final convergence.
\section{Experiments}
\subsection{Simulation study}
We define our canonical projection geometry following the intrinsic parameter of a Siemens CIOS Fusion C-Arm, which has image dimensions of $1536\times 1536$, isotropic pixel spacing of 0.194~mm/pixel, a source-to-detector distance of 1020~mm. We downsample the detector dimension to be $128\times 128$. We trained our algorithm using 17 full body CT scans from the NIH Cancer Imaging Archive~\cite{roth2014new} and left 1 CT for testing. The pelvis bone is segmented using an automatic method in~\cite{krvcah2011fully}. CTs and segmentations are cropped to the pelvis cubic region and downsampled to the size of $128\times 128\times 128$. The world coordinate frame origin is set at center of the processed volume, which is 400~mm above the detector plane center.\\
At training iteration $i$, we randomly sample a pair of pose parameters, $(\theta^i, \theta_t^i)$, rotation from $N(0, 20)$ in degree, translation from $N(0, 37.5)$ in mm, in all three axes. We then randomly select a CT and its segmentation, $V_{CT}$ and $V_{Seg}$. The target image is generated online from $V_{CT}$ and $\theta_t^i$ using our ProST. $V_{Seg}$ and $\theta$ are used as input to our network forward pass. The network is trained using SGD optimizer with a cyclic learning rate between 1e-6 and 1e-4 every 100 steps~\cite{smith2017cyclical} and a momentum of 0.9. Batch size is chosen as 2 and we trained 100k iterations until convergence.\\
We performed the 2D/3D registration application by randomly choosing a pose pair from the same training distribution, ${(\theta^R, \theta_t^R})$. Target image is generated from the testing CT and $\theta_t^R$. We then use SGD optimizer to optimize over $\theta^R$ with a learning rate of 0.01, momentum of 0.9 for iteration. We calculate the STD of the last 10 iterations of $L_N$ as $std_{L_N}$, and set a stop criterion of $std_{L_N}<3\times 10^{-3}$, then we switch to Gradient-NCC similarity using SGD optimizer with cyclic learning rate between 1e-3 and 3e-3, and set the stop criterion, $std_{L_{NCC}}<1\times 10^{-5}$. We conduct in total of 150 simulation studies for testing our algorithm.

\subsection{Real X-ray study} 
We collected 10 real X-ray images from a cadaver specimen. Groundtruth pose is obtained by injecting metallic BBs with 1~mm diameter into the surface of the bone and manually annotated from the X-ray images and CT scan. The pose is recovered by solving a PnP problem~\cite{hartley2003multiple}. For each X-ray image, we randomly choose a pose parameter, rotation from $N(0, 15)$ in degree, translation from $N(0, 30)$ in mm, in all three axes. 10 registrations are performed for each image using the same pipeline, resulting in a total of 100 registrations.   

\begin{figure}[h!]
\includegraphics[width=\textwidth]{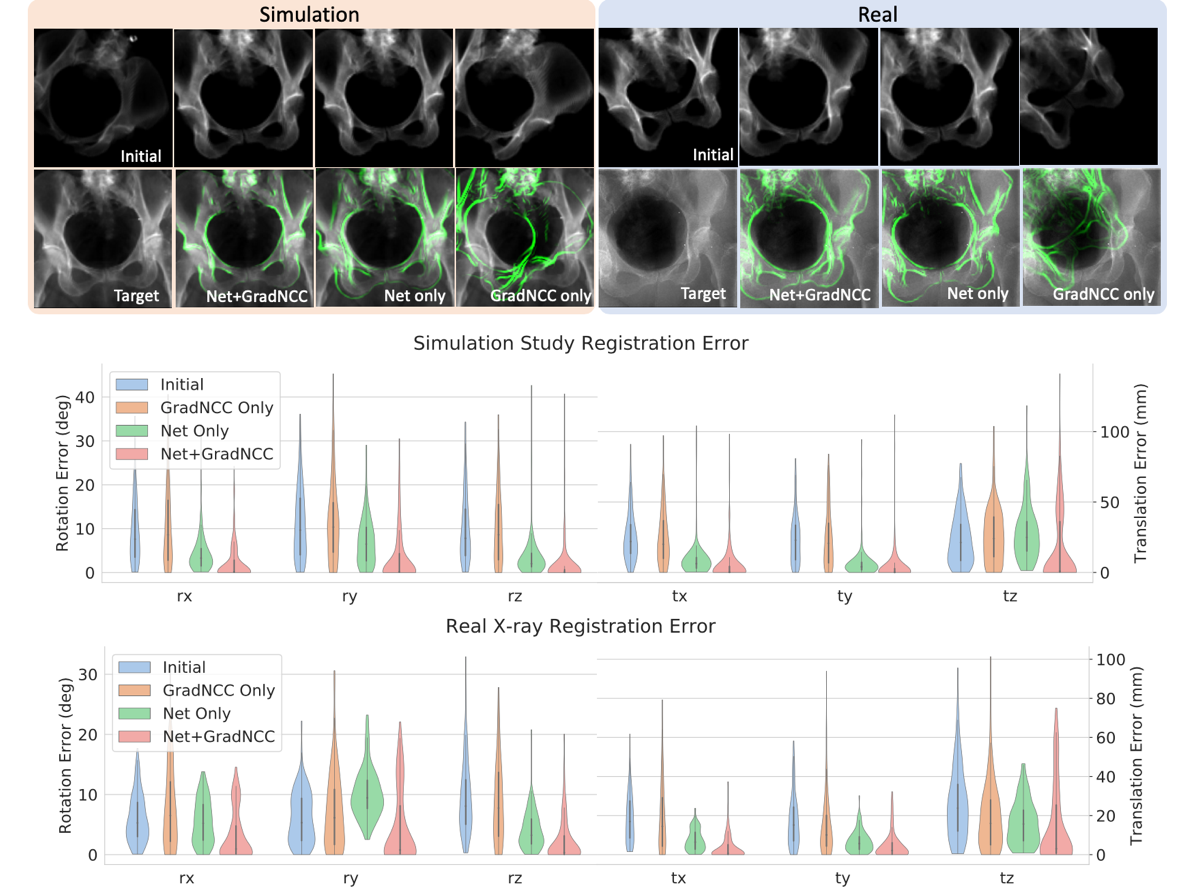}
\caption{The top row shows qualitative examples of Net+GradNCC, Net only, GradNCC only convergence overlap, for simulation and real X-ray respectively. The middle is the registration error distribution of simulation. The bottom is the distribution for real X-ray experiments. x, y and z-axis correspond to LR, IS and AP views.} \label{fig3:result}
\end{figure}
\section{Results}
\begin{table}
\centering
\label{table} 
\begin{tabular}{ |c|c|c|c|c|c| } 
\hline
\multicolumn{2}{|c|}{} & \multicolumn{2}{|c|}{Simulation Study} & \multicolumn{2}{|c|}{Real X-ray Study}\\
\cline{3-6}
\multicolumn{2}{|c|}{} & Translation & Rotation & Translation & Rotation \\
\hline
\multicolumn{2}{|c|}{Initialization} &  $41.57 \pm 18.01$ & $21.16\pm 9.27$ & $30.50\pm 13.90$ & $14.22\pm 5.56$\\
\hline
           \multirow{2}{*}{GradNCC} & mean  & $41.83\pm23.08$  & $21.97\pm11.26$ & $29.52\pm20.51$ & $15.76\pm8.37$\\ \cline{2-6}
          & median & 38.30  & 22.12 & 26.28 & 16.35 \\   
\hline
         \multirow{2}{*}{Net}  & mean  & $13.10\pm18.53$ & $10.21\pm7.55$  & $12.14\pm6.44$ & $13.00\pm4.42$ \\ \cline{2-6}
            & median &  9.85 &  9.47 & 11.06 & 12.61\\  
\hline
Net+& mean   & $7.83\pm19.8$  & $4.94\pm8.78$ & $7.02\pm9.22$ & $6.94\pm7.47$\\ \cline{2-6}
GradNCC & median & 0.25  &  0.27 & 2.89 & 3.76\\   
\hline
\end{tabular}
\vspace{0.5em}
\caption {Quantitative Results of 2D/3D Registration}
\end{table}

We compared the performance of three methods, which are Grad-NCC only, Net only, and Net+GradNCC. The registration accuracy was used as the evaluation metric, where the rotation and translation errors are expressed in degree and millimeter, respectively. The coordinate frame $F^r$ are used to define the origin and orientation of the pose. In Fig.~\ref{fig3:result}, both qualitative and quantitative results on the testing data are shown. Numeric results are shown in Table.~\ref{table}. The Net+GradNCC works the best among comparisons in both studies.

\section{Discussion}
We have seen from the results that our method largely increases the capture range of 2D/3D registration. Our method follows the same iterative optimization design as the intensity-based registration methods, where the only difference is that we take advantage of the great expressivity of deep network to learn a set of more complicated filters than the conventional hand-crafted ones. This potentially makes generalization easier because the mapping that our method needs to learn is simple. In the experiment, we observed that the translation along the depth direction is less accurate than other directions in both simulation and real studies, as shown in Fig.~\ref{fig3:result}, which we attribute to the current design of the network architecture and will work on that as a future direction.

\section{Conclusion}
We propose a novel Projective Spatial Transformer module (ProST) that generalizes spatial transformers to projective geometry, which enables differentiable volume rendering. We apply this to an example application of 2D/3D registration between radiographs and CT scans with an end-to-end learning architecture that approximates convex loss function. We believe this is the first time that spatial transformers have been introduced for projective geometry and our developments will benefit related 3D research applications. 

\bibliography{ref}
\bibliographystyle{splncs04}
\end{document}